\documentclass[letterpaper, 10 pt, conference]{ieeeconf}
\IEEEoverridecommandlockouts
\usepackage{comment}
\usepackage{url}
\usepackage{amsmath}
\usepackage{bm}
\usepackage{todonotes}
\usepackage{amssymb}
\usepackage{acronym}
\usepackage{color}
\usepackage{algorithm}
\usepackage{algorithmic}
\usepackage{nomencl}
\usepackage{caption}
\usepackage{subcaption}
\makenomenclature

\usepackage{amsthm}

\usepackage[utf8]{inputenc}
\usepackage[english]{babel}
 
\newtheorem{definition}{Definition}

\newtheorem{problem}{Problem}
\newtheorem{example}{Example}

\newtheorem{lemma}{Lemma}

\newtheorem{remark}{Remark}

\newif\ifuseboldmathops
\newif\ifuseittextabbrevs
\useboldmathopstrue   % comment out to use mathbb
\ifuseittextabbrevs

	\newcommand{\ie}{{\it i.e.}}

\else

	\newcommand{\ie}{i.e.}

\fi

\ifuseboldmathops
	\newcommand{\reals}{\mathbf{R}}

\else
	\newcommand{\reals}{\mathbb{R}}

\fi

\ifuseboldmathops

\else

\fi

\ifuseboldmathops

\else

\fi

\ifuseboldmathops

\else

\fi

\newcommand{\parent}{\mathsf{Parent}}

\newcommand{\abs}[1]{\lvert#1\rvert}

\newcommand{\indices}{I}
\acrodef{mdp}[MDP]{Markov Decision Process}
\acrodef{pomdp}[POMDP]{Partially Observable Markov Decision Process}

\newcommand{\calF}{\mathcal{F}}
\newcommand{\calM}{\mathcal{M}}

\newcommand{\dist}{\mathcal{D}}

\newcommand{\supp}{\mathsf{Supp}}

\usepackage{accents}

\acrodef{dbn}[DBN]{dynamic Bayesian network}
\acrodef{scltl}[sc-LTL]{syntactically co-safe LTL}
\acrodef{ltl}[ltl]{Linear temporal logic}
\acrodef{dfa}[DFA]{deterministic finite-state automaton}
\acrodef{ssp}[SSP]{Stochastic Shortest Path}

\usepackage{acronym}

\begin{document}
\title{Attention-Based Probabilistic Planning with Active Perception}

%\author{\IEEEauthorblockN{Haoxiang Ma}
%\IEEEauthorblockA{Department of Robotic Engineering\\
%Worcester Polytechnic Institute\\
%Worcester, Massachusetts 01609\\
%Email: hma2@wpi.edu}
%\and
%\IEEEauthorblockN{Jie Fu}
%\IEEEauthorblockA{Department of Robotic Engineering\\
%Worcester Polytechnic Institute\\
%Worcester, Massachusetts 01609\\
%Email: jfu2@wpi.edu}}

\author{Haoxiang Ma, Jie Fu
% <-this % stops a space
 \thanks{H. Ma and J. Fu are with the Department
 of Robotics Engineering, Worcester Polytechnic Institute, Worcester, MA,  USA, 01609  e-mail: hma2, jfu2@wpi.edu}}

\maketitle

\begin{abstract}
Attention control is a key cognitive ability for humans to select information relevant to the current task. This paper develops a computational model  of attention and an algorithm for attention-based probabilistic planning in Markov decision processes. In attention-based planning, the robot decides to be in different attention modes. An attention mode corresponds to a subset of state variables monitored by the robot. By switching between different attention modes, the robot actively perceives task-relevant information to reduce the cost of information acquisition and processing,  while achieving near-optimal task performance. Though planning with attention-based active perception inevitably introduces partial observations, a partially observable MDP formulation makes the problem computational expensive to solve. Instead,  our proposed method employs a hierarchical planning framework in which the robot determines what to pay attention to and for how long the attention should be sustained before shifting to other information sources. During the attention sustaining phase, the robot carries out a sub-policy, computed from an abstraction of the original MDP given the current attention.  We use an example where a robot is tasked to capture a set of intruders in a stochastic gridworld. The experimental results show that the proposed method enables information- and computation-efficient optimal planning in stochastic environments.  \end{abstract}

% % Note that keywords are not normally used for peerreview papers.
% \begin{IEEEkeywords}
% Markov decision processes, active 
%  \end{IEEEkeywords}

% \fi
\IEEEpeerreviewmaketitle
\section{Introduction}

Attention is a core component in human's perceptual and cognitive functions. With limited cognitive resources, a person only focuses  on a subset of information and/or a subset of tasks for a period of time. Despite such cognitive limits, humans excel complex decision-making problems, enabled by the  attention control  mechanism   that answers, where to allocate attention? When to switch the attention? And how to coordinate attention and actions given the task?
It is desirable to equip autonomous robots with such human-like cognitive flexibility, for reducing the cost of computation, communication, and information processing when performing tasks in uncertain environments. This paper develops a computational model of attention and an attention-based probabilistic planning in \ac{mdp} that balances optimal task performance and the cost of information acquisition.  
% Using the computational model, we develop

%The autonomous attention-based planning method answers: Where to allocate attention given a set of information sources and the current task performance? When to switch attention from one aspect of the information to anther? How to coordinate attention and actions when planning in an uncertain environment?
 
 The proposed computational model of attention is inspired   by spotlight attention--a class of ``top down'' attention that allows a person to flexibly select one subset of information to process at the expense of the others  \cite{buschman2010shifting}. In neuron-psychology, researchers have shown that goal-directed behavior requires a person to focus on a subset of particular stimuli despite the fact that they might not be the most salient, and switch between different subsets of stimuli given the task progress. This process, called selective attention \cite{johnston1986selective}, is the differential processing of multiple simultaneous sources of information.
 Spotlight attention, combined with selective attention, are fundamentals to cognitive control \cite{hanania2010selective,lee2013critical}. 
 
  We thus investigate the role of attention for probabilistic planning, modeled  an \ac{mdp} with a factored state   representation. We introduce spotlight and selective attention  as the process of  active state information acquisition. An attention mode is  characterized by an information acquisition and processing function used by the agent to select a subset of state variables to monitor and process at a time. Different choices of information sources constitute different attention modes  which the agent can choose to be in.
   With attention modes incorporated, the agent  plans two levels of decisions: At the high level, the agent selects an attention mode--what to pay attention to, and for how long it will sustain that attention mode; at the low level, the agent takes actions given the information being attended to, \ie, acquired and processed,  given its attention mode. In particular, the agent takes low-level actions by following a subpolicy, which is optimal in the abstraction of the original \ac{mdp} given its current attention mode. Each state in the abstraction aggregates a set of states in the original \ac{mdp} that are observation-equivalent given the choice of the attention mode.  The low-level subpolicies do not require information other than these provided by the current attention mode. After following the subpolicy for a pre-determined number of steps,  the agent  observes the full state information, and decides if it needs to switch its attention and subpolicy. The hierarchical planning given attention control employs multi-objective probabilistic planning that  trades off task performance measured by the total discounted rewards and the cost of information acquisition and processing. 
 
 \subsection{Related Work}  Attention models have been introduced in machine learning \cite{velivckovic2018graph},  reinforcement learning \cite{zambaldi2018relational,huang2020inner},  and machine translation \cite{vaswani2017attention}.  
 A common feature of these work is to introduce an attention mechanism that  selects relevant features to be processed from the data. These attention models are  embedded into the structure of neural networks as either a blackbox or a greybox and  implicitly learned from data. %In \cite{huang2020inner}, the authors look into the  application of heterogeneous multi-robot systems in complex scenarios, develop an inner attention method to accurately allocate limited team capacity to satisfy different tasks. The inner attention mechanism is designed based on the attention mechanism and a multi-agent reinforcement learning algorithm.
 Differs from these work, our model of attention is explicit and the agent employs optimization-based decision-making to decide which information is attended to and justify the reasoning.
 
 Our work is also related to  active perception \cite{spaan2015decision,satsangiExploitingSubmodularValue2018,spaan2010active,spaan2008cooperative}. In an active perception task, an agent  selects sensory actions to reduce its uncertainty about one or more hidden variables for tasks such as surveilance and multi-people tracking. The solution of partially observable MDPs (POMDPs) is employed for planning an active perception strategy where the  action space is a set of sensors to be selected from and the reward function is chosen to be inverse to the uncertainty about the hidden state such as the entropy of the belief. In recent work \cite{satsangiExploitingSubmodularValue2018}, the authors established  the condition when the value function in active perception is submodular. The submodularity property is used to reduce the computational cost of solving the formulated POMDPs. Differ from active perception, our work focuses on the joint active perception and task planning where the perception and attention is task-oriented. Our problem is closely related to  \cite{ghasemi2019online}, where the authors studied an online active perception and task planning  where the agent is to complete a task captured by a reward function, with  limited information gathering budget. 
 They also employed submodularity in information gathering to design an approximate-optimal greedy sensor selection strategy. In addition, they show that a reduction in the entropy of belief leads to better future reward due to the convexity of value functions w.r.t. beliefs in POMDPs.  While it is natural to use POMDP models for  attention-based planning, the reduction in the cost of information acquisition comes with  computational burden, as POMDPs are NP-hard \cite{lee2008makes}. Our approach uses abstraction-based planning and hierarchical planning to mitigate the need of POMDP planning despite the partial observations introduced by active information acquisition. 

The rest of the paper is organized as follows: We introduce preliminaries and formalize the problem in Section~\ref{sec: preliminaries}. The modeling approach and solution algorithms are presented in section~\ref{sec: plan}. Section~\ref{sec: experiment} uses an example to illustrate the effectiveness of the method. Section~\ref{sec: conclusion} concludes.

\section{Preliminaries and Problem formulation}
\label{sec: preliminaries}
Notations: The set of real $n$-vector  is denoted $\reals^n$. $\reals^+$ is a set of nonnegative real numbers. The powerset of a given set $\indices$ is denoted by $2^{\indices}$. The notation $z_i$ refers to the $i$-th component of vector $z\in \reals^n$, or to the $i$-th element of a set or a sequence  $z_1 , z_2 ,\ldots$, which will be made clear given the context. Given a finite, discrete set $Z$, the set of possible distributions over $Z$ is denoted $\dist(Z)$. Given a distribution $d \in \dist(Z)$, we denote $\supp(d) = \{z\in Z\mid d(z)>0\}$ be the support of the distribution $d$.
 %\printnomenclature

\subsection{Preliminaries}

 We consider a stochastic system modeled as an  \ac{mdp}  which has factored state space representation (but not necessarily a factored transition function as in factored \ac{mdp}s \cite{guestrin2003efficient}). In the \ac{mdp} $M = (\bm{X}, A, P, x_0, \gamma, R)$, the state is described via a set of random variables $X = [X_1, \ldots,X_n]$, where each variable $X_i$ can take a value in a finite domain $\bm{X}_i$ and $\bm{X}$ is the set of states.
%  $ X = [X_1,\ldots,X_n]$ where each variable $X_i$ can take a value in a finite domain $\bm{X}_i$. We use $\indices=\{1,\ldots, n\}$ to denote the indices of random variables in the factored \ac{mdp} state.
% The factored transition function is defined as follows: For each action $a\in A$, the transition function given $a$ is represented using a \ac{dbn}\cite{dean1989model}. Let $ X$ be the state variable at the current time step and $ X'$ be the variable at the next time step. The transition graph of the DBN contains nodes $\{X_1, \ldots, X_n, X_1',\ldots, X_n'\}$ and the parents of $X_i'$ in the graph is denoted by $\parent_a(X_i')$ for action $a$. Each node  $X_i'$ is associated with a conditional probability $P(X_i' | \parent_a(X_i'), a)$.
% The transition probability function $P$ can be factored as
% \[
% P(x' \mid x, a) = \prod\limits_{i=1}^n P( x'_i|\vec{z}_i, a ),
% \]
% where $\vec{z}_i$ is the value in $x$ of the variables in $\parent_a(X_i)$.
The reward function of the  \ac{mdp} is defined as usual: $R:\bm{X}\times A\rightarrow \reals$ such that $R(x, a)$ represents the reward obtained by the agent at state $x$ after taking action $a$. $P: \bm{X}\times A\rightarrow \dist(\bm{X})$  is the probabilistic transition function. Let $x_0 \in \bm{X}$ represent the initial state. 
We consider the \ac{mdp} has an infinite horizon and future rewards can be discounted with discounting factor $\gamma \in (0,1]$. When $\gamma=1$, the future reward is not discounted.

A Markovian, randomized policy in the  \ac{mdp} $M$ is a function $\pi: \bm{X} \rightarrow \dist(A)$. 
Each policy is associated with a value function $V^\pi$ where $V^\pi (x)$ is the future (discounted) reward starting from $x$ and following policy $\pi$. 
\[
V^\pi(x)=\mathbb{E}\left[\sum_{t=0}^\infty \gamma^t R(x_t,\pi(x_t)\mid x_0=x)\right].
\]

The optimal Markov policy in $M$ is denoted $\pi^{\ast}$ and the optimal value function is $V^{\ast}$ that satisfies the Bellman optimality equation:
\[
V^{\ast}(x) = \max_{a \in A}(R(x, a) + \sum_{x' \in \bm{X}}P(x'|x, a)V^{\ast}(x')), \forall x \in \bm{X}.
\]

Given human's limited information processing capabilities, the attention control mechanism enables a person to select a subset of information to process at a time.  We introduce  the \emph{cost of  information acquisition} as a quantitative measure of the cognitive workload for acquiring and processing information. 

\begin{definition}[Cost of information acquisition]
For each variable $X_i$ in the  multivariate random variable $ X$, the cost of acquiring the value of $X_i$ is given by $c_i \in \reals^+$.  
\end{definition}

Now, we state the problem informally:
\begin{problem}
\label{problem}
Given  an \ac{mdp} $M = (\bm{X}, A, P, x_0,\gamma, R)$, determine an optimal policy  that balances the two objectives: 1) maximizing the total discounted reward in the  \ac{mdp}; and 2) minimizing the total discounted cost of information acquisition.
\end{problem}

\section{Attention-based Active Perception and Probabilistic Planning}
\label{sec: plan}
\subsection{Abstraction with Spotlight Attention}
The spotlight model of attention \cite{buschman2010shifting} is a model of human's visual attention in analogy of spotlight. Information outside of this spotlight is assumed to be overlooked (or not be attended to). We introduce a computational model of spotlight attention as part of our solutions for information-efficient  probabilistic planning with active perception.

\begin{definition}[A spotlight attention function]
Given the indices $\indices= \{1,\ldots, n\}$ of variables in an \ac{mdp} $M$. Let $\mathcal{I}=\{I_0,I_1,\ldots, I_m\}$ with $m < \vert{2^\indices}\vert$ be a subset of $2^{\indices}$ and  $I_k \in \mathcal{I}$ be the $k$-th element in the set. A \emph{spotlight attention function under mode} $k$, denoted  $f_k: \bm{X}  \rightarrow \bm{Y}$ is defined such that $f_k(x)$ maps a state vector $x$ to a subvector $y \in \bm{Y}$ containing the components in $x$ whose indices $i$ are in $I_k$. Formally, $f_k(x)=[x_i]_{i \in I_k}$.
\end{definition}
Note that $f_k$ is a surjective mapping. The inverse is defined by $f_k^{-1}(y)= \{ x\mid f_k(x)=y\}$. In general, the total number of attention modes can be up to $\abs{2^\indices}$. But in practice, a robot may have limited information processing capabilities and have a smaller number of attention modes. We define $I_0 = \indices$ and the corresponding attention function $f_0$ is the null attention function, by using which every state variable is observed by the robot. 

When the robot uses an attention function $f_k$, at every state, it reduces the cost of information acquisition by $ C_{k}= \sum_{j\notin I_k} c_j$ for not acquiring the information about unattended state variables. We refer to this reduction in the cost as the \emph{sensor deactivating reward}.

The spotlight attention introduces partial observations for the robot. To avoid the POMDP formulation, we   consider that  a spotlight attention would allow the robot to view the stochastic system through a lens of abstraction: If a subset of states provide the same observation under one attention mode, then these states are aggregated. Such an aggregation is formalized through two definitions: A disaggregation probability that shows how an aggregated state represents the actual state in the \ac{mdp}, and an abstract \ac{mdp} constructed from state-aggregation in the original \ac{mdp}.

\begin{definition}[Disaggregation probability  extended from \cite{bertsekas2018feature,hutter2014extreme}]
Given an attention function $f_k$, a disaggregation probability  function $D_k: \bm{Y} \rightarrow \dist(\bm{X})$ maps an observed state $y$ to a distribution $D_k(\cdot | y)$ where $D_k(x|y)$ is the probability of $x$ given state $y$ and satisfies the following constraint, 
\[
D_k(x|y) = 0 \text{ if  } f_k(x) \ne y. 
\]
\end{definition}
The constraint means that  the function $D_k$  assigns zero probability to any state $x \notin f_k^{-1}(y)$.
By definition, there can be more than one disaggregation probability function that maps an observed state into a distribution over $\bm{X}$. 

\begin{definition}[Attentional \ac{mdp} with state-aggregation]
Given an \ac{mdp} $M = (\bm{X}, A, P, x_0, \gamma, R)$, a disaggregation probability function $D_k:\bm{Y}\rightarrow \dist(\bm{X})$, and an attention function $f_k: \bm{X}  \rightarrow \bm{Y}$, an \emph{attentional MDP} respecting $f_k$ and $D_k$ is 
\[
M_k = (\bm{Y}, A, P_k, y_0, \gamma, r_k),
\]
where $y_0 = f_k(x_0)$, and the probabilistic transition function and reward function are obtained by marginalization as follows,
\[
P_k(y ' \mid y,  a) = \sum_{x'\in f_k^{-1}(y')} P(x'|x,a) D_k(x|y).
\]
\[
r_k(y,a) = \sum_{x\in f_k^{-1}(y)} R(x,a)D_k(x|y).
\]
\end{definition}
The disaggregation probabibility can be understood as how representative the aggregated state $y$ is for a state $x$ being aggregated by $f_k$ to $y$. It can be determined to minimize the approximation error between the approximate  optimal value/policy functions, solved from the abstraction, herein $M_k$,  and the optimal value/policy functions in the original \ac{mdp} $M$ (see detailed discussions in \cite{bertsekas2018feature}). In this paper, we use the notion of state aggregation to capture the models obtained with different spotlight attention functions. As minimizing approximation error is not our key focus, uniform distributions over observation-equivalent states can be used in defining a disaggregation probability function.
%  as our focus is not to reduce the sub-optimality in the solution obtained by abstraction-based planning.  

%In words, the attention function allows us to focus on an MDP with a smaller state space. 

By solving the attentional \ac{mdp} $M_k$ for each  $k=0,\ldots, m$, we obtain  an optimal Markov policy  $\pi_k: \bm{Y}\rightarrow \dist(A)$, referred to as the \emph{attention-based policy in mode $k$}.  Note that since the null attention function $f_0$ is included, $M_0$ is exactly the same as the original  \ac{mdp} $M$.

\begin{remark}
As a remark, we discuss a special case of \ac{mdp} and ways to construct the set of attention functions and abstract attentional \ac{mdp}s. In a factored \ac{mdp} \cite{guestrin2003efficient}, not only the state is of factored representation but also the transition function is factored. 
 The factored transition function is defined as follows: For each action $a\in A$, the transition function given $a$ is represented using a \ac{dbn}\cite{dean1989model}. Let $ X$ be the state variable at the current time step and $ X'$ be the variable at the next time step. The transition graph of the DBN contains nodes $\{X_1, \ldots, X_n, X_1',\ldots, X_n'\}$ and the parents of $X_i'$ in the graph is denoted by $\parent_a(X_i')$ for action $a$. Each node  $X_i'$ is associated with a conditional probability $P(X_i' | \parent_a(X_i'), a)$.
  The probabilistic transition  function $P$ can be factored as
\[
P(x' \mid x, a) = \prod\limits_{i=1}^n P( x'_i|\vec{z}_i, a ),
\]
where $\vec{z}_i$ is the value in $x$ of the variables in $\parent_a(X_i)$. In this case, we may define an attention mode to monitor a state variable as well as its parents. By doing so, the probabilistic transition function in the attentional \ac{mdp} can be obtained from the Bayesian network directly. For example, let $I_k$ be a set of indices of variables where if $i\in I_k$, then its parents for any actions are in $I_k$ as well. The following transition function is well-defined: 
\[
P_k(y' \mid y, a) = \prod\limits_{i \in I_k} P( x'_i|\vec{z}_i, a )\]
where $\vec{z}_i$ is the value in $y$ of the variables in $\parent_a(X_i)$.
\end{remark}

% The idea for this reward definition is that when the attention only focuses on a subset of variables, the agent is either unaware of the values of other variables or does not track the evolution of the values of other variables (belief). When the agent is optimistic (resp. pessimistic), he thinks that he will get the maximal (resp. minimal) of rewards that could be obtained within the set of observation equivalent states. 

\subsection{Information-efficient   planning with attention control}
Given that different attention functions spotlight on different subsets of state variables, the optimal policy $\pi_k$ for $k\ne 0$  may not be optimal in the original \ac{mdp}. We consider the robot can switch attentions and use different optimal attentional policies $\{\pi_k,k=1,\ldots, m\}$ from time to time. Note that the null attentional policy $\pi_0$ are excluded as it is the original \ac{mdp}. The optimal switching between subpolicies can be modeled as a semi-MDP \cite{sutton1999between} where the subpolices $\Pi= \{\pi_k, k=1,\ldots, m\}$ are options.

To solve Problem~\ref{problem}, we propose an attention-shift \ac{mdp} framework, which enables the planning robot to actively switching between different attention functions for simultaneously optimizing
the total (discounted) rewards and reducing the cost of information acquisition. 
\begin{definition}

Given the   MDP $M = (\bm{X},A,P, x_0, \gamma, R)$, a set $\Pi = \{\pi_k \mid k= 1, \ldots, m\}$ of   attention-based sub-policies, a time bound $T\ge 1$, the \emph{attention-shift \ac{mdp}} is a tuple 
\[
{\cal M}_T = (\bm{X}, \Pi \times  \{1,\ldots, T\}, \hat P_T,x_0, \gamma, R_T^G, R_T^I),
\]
with the following components:
\begin{itemize}
    \item $\bm{X}$ is a finite set of states in the original \ac{mdp}.
    \item $\Pi\times \{1,\ldots, T\}$ is a set of \emph{sustained attention selection actions}. The first component in this action is an attention-based policy in $\Pi$, the second component describes how many steps this policy should be applied (or sustained) before shifting to another policy (or attention). We refer the second component as the \emph{attention sustaining time}. $T$ is an upper bound on attention sustaining time.
    \item   $\hat P_T$: $\bm{X} \times (\Pi \times  \{1,\ldots, T\}) \rightarrow D(\bm{X})$ is the probability transition function, defined as follows: Given state $x$ and selection $(\pi_k,t)$,
    \[
        \hat P_T(x'\mid x,(\pi_k, t)) = P^t(x'|x, \pi_k),
    \]
    where $P^t(x'|x,\pi_k)$ is the $t$-step transition probability from $x$ to $x'$ in  the Markov chain induced by $\pi_k$ from the   MDP $M$.
    \item $R_{T}^G$: $\bm{X} \times(\Pi \times \{1, \ldots, T\}) \rightarrow \mathbf{R}$ is the goal-directed reward function, defined as follows: Given state $x$  and action $(\pi_k, t)$,
    \[
     R_T^G(x, (\pi_k, t)) =E_{\pi_k}[\sum_{j=1}^{t} \gamma^{j-1}r_j], \] %r(s_t,\pi^\ast_i).
    % \]
    % where $r(s_t,\pi^\ast_i) = 
    which is the   discounted total reward when the policy $\pi_k$ is applied to the original  MDP for a duration of $t$ time steps starting from the state $x$. This can be obtained from evaluating policy $\pi_k$ in $M$.  Here, $r_j$ is the reward obtained by following $\pi_k$ at time step $j$.
%    \todo{Add}.
    \item $R_{T}^I$: $\Pi \times \{1, \ldots, T\} \rightarrow \mathbf{R}^+$ is the sensor deactivating rewards and only depends on the attention-based policy. It is  
    defined as follows: Give an action $(\pi_k, t)$,
    \[
        R_{T}^I(\pi_k, t) = \sum_{j = 1}^t\gamma^{j-1}  C_k,     \]
    where  $C_k= \left(\sum_{j \notin I_k} c_j \right) $ is one-step sensor deactivating reward  based on the current attention function $f_k$. 
    \end{itemize} 
\end{definition}

 In this attention shift \ac{mdp}, suppose the robot selects $\pi_k $ for $t$ steps, then it will carry out the policy which only pays attention to the states in $I_k$ for $t$ steps. After the $t$-th step, the robot will observe the value of the current state in the \ac{mdp}, determine the next attentional policy and how long that policy should be sustained. 
 
%  Considering the agent is to plan a Pareto-optimal policy that trades off between two objectives: 1) Maximizing the total discounted goal-directed rewards given reward function $R^G_T$; and 2) Maximizing the total discounted sensor deactivating rewards given the reward function $R^I_T$. The second objective is equivalent to minimizing the total discounted cost of information acquisition.
 An approximate optimal solution to Problem~\ref{problem} can be reduced to Pareto-optimal planning in the attention shift \ac{mdp} given the multiple objectives: maximizing the total reward and miminizing the cost of information acquisition. We employ linear scalarization method for multi-objective planning. Note that more advanced multi-objective planning methods can be used (see a survey in \cite{roijers2013survey}).
 \begin{definition}[Weighted Sum Reward Function]
 Given a weighting parameter $w = [w_1, w_2]$, we define the weighted sum reward function as:
 \[
 R^w_T(x,(\pi_k, t)) = (w_1 R^G_T(x,(\pi_k, t))  
     + w_2 R^I_T(\pi_k, t))
 \]
 where $w_1>0,w_2 >0$ are the weights assigned to two different objectives and $w_1+w_2=1$. 
 \end{definition}
 The robot's target is finding the Pareto-optimal policy by solving the value function that satisfies the following Bellman  equation:
\begin{align}
    \begin{split}
    \hat V(x) =
    & \max_{(\pi_k, t) \in \Pi \times  \{1,\ldots, T\}}
    R^w_T(x,(\pi_k, t)) \\ 
    + &  \gamma^t\sum_{x'}\hat V(x')\hat P_T(x'\mid x,(\pi_k, t)), \forall x\in \bm{X}.
    \end{split}
\end{align}
 
The maximal time span $T$ is a parameter in constructing the \ac{mdp} with attention shift. Let $\hat V_T^\ast$ be the optimal value function of the optimal policy $\hat \pi^{\ast}_T$ in ${\cal M}_T$. We can prove the following Lemma. 

\begin{lemma}
\label{nosensorreward}
If $R_T^I(\pi_k, t) = 0,  \forall (\pi, t) \in \Pi \times \{1, \ldots, T\}$, that is, there is no sensor deactivating rewards, then $\hat V_1^\ast= \hat V_T^\ast, \forall T > 1$.
\end{lemma}

\begin{proof}
The two \ac{mdp}s $\calM_1$ and $\calM_T$ for any $T > 1$ share the same state space and the same reward function as the original \ac{mdp}. If we project the action sets of $\calM_T$ to that action set of $\calM_1$, we obtain the same \ac{mdp} as $\calM_1$. 

Now, give the optimal policy in $\calM_T$, we can show that this optimal policy is equivalent to a finite-memory policy in $\calM_1$ as follows, suppose that $
\hat \pi^\ast_T(x) = (\pi_k,t)
$, then it is equivalent to the action sequence of $(\pi_k,1)$ repeated for $t$ time steps. The memory state is to keep track of the number of repetition. 

It is known that in an \ac{mdp} with total discounted rewards, the optimal value can be attained by a Markov policy \cite{puterman2014markov}. Thus, the value $\hat V_1^\ast \ge \hat V_T^\ast$ where $\hat V_T^\ast$ equals the value of the finite-memory policy $\hat \pi^\ast_T$ evaluated in the $\calM_1$. On the other hand, because $\calM_1$ and $\calM_T$ share the same state space, the actions of $\calM_1$  is a subset of actions of $\calM_T$, the transitions given actions in $\calM_1$ are the same for both $\calM_1$ and $\calM_T$,  we then have the optimal value $\hat V_T^\ast \ge \hat V_1^\ast$. Combining the two inequality, we have $\hat V_1^\ast   = \hat V_T^\ast$.   
\end{proof}

This lemma tells us that if there is no cost of information acquisition, then the robot can find the optimal strategy by  allowing the  attention switching given \emph{complete observation of state variables at every time step}.

However, if we introduce the cost of information acquisition, then the robot may prefer to observe a subset of information sources using spotlight attention and sustain this spotlight attention for a period of time. Because during the attention sustaining period, the robot saves the cost of acquiring information for these unattended state variables.  For example, if $(\pi_1, 10)$
is selected, then the robot will carry out the policy following $f_1(\bm{X})$ for 10 time steps. 

The question is, in search of the optimal attention shift policy, how to select the parameter $T$ so that for a given weighting parameter, the value $\hat V_T^\ast = \hat  V_{T'}^\ast$ for any $T' > T$. 

\begin{definition} 
A discrete number $T$ is called the \emph{optimal bound on attention sustaining} if for any $T' >T$, $\hat V_T ^\ast = \hat V_{T'}^\ast$ \footnote{The equality is compared element-wise.}. 
\end{definition}

One naive approach is to construct the attention-shift \ac{mdp} for a large $T$, and check the solved optimal policy to determine the longest attention sustaining time spans at every state. If the longest attention sustaining time is $T$, then the robot can increase $T$. If it is less than $T$, then the optimal parameter $T$ should be the longest attention sustaining time. 

However, this naive approach requires to solve an \ac{mdp} with a large action set (with a large enough $T$). It is easy to prove that for any $T_1, T_2 \in \mathbb{N}$, if $T_1 > T_2$, then $\hat V_{T_1}^\ast \ge \hat V_{T_2}^\ast$. 
The claim is a generalization of Lemma~\ref{nosensorreward} as the \ac{mdp} $\calM_{T_1}$ is a sub-MDP of $\calM_{T_2}$.

Thus, another approach is to increase $T$ and solve a sequence of attention shift \ac{mdp}s with increasing $T$ until a bound is reached. This second approach may use the value $\hat V^\ast_T$ as a lower bound and initialization for value iteration for the optimal value $\hat V^\ast_{T+1}$ to improve the convergence.

The complexity of value iteration is $\mathcal{O}(\vert\bm{X}\vert^2\vert A\vert)$. In ${\cal M}_T,$ the size of the state space is $\vert \bm{X} \vert$, the size of the action space is $\vert \Pi\vert \cdot T$.  When solving attention-based sub-policies, the complexity of value iteration is $\mathcal{O}(|\bm{Y}|^2|A|)$ where $\bm{Y}$ is the state space of $M_k$, for $k = 1, 2, \cdots, m$. Since $\vert \bm{Y} \vert \ll \vert \bm{X} \vert$, the whole computation complexity should be $\mathcal{O}(\vert \bm{X} \vert ^2 \vert\Pi\vert T)$. Comparing to a POMDP formulation, our approach  reduces the computation complexity at a cost of optimality due to the use of semi-\ac{mdp}  and abstraction-based planning.

\section{Experiments}
\label{sec: experiment}
\begin{figure}
    \centering
    \includegraphics[width = 0.35\linewidth]{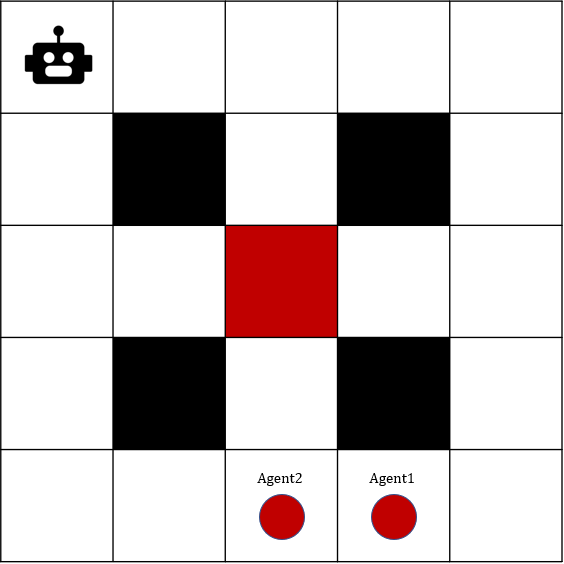}
    \caption{The gridworld with two dynamic agents. Black cells represent walls, red cell represents the penalty cell. The two dots are agents.}
    \label{image_map}
\end{figure}

We consider an example of  robot motion planning in stochastic environment, as shown in Figure~\ref{image_map}. The robot is assigned with the task of capturing two stochastic agents. Once both agents are captured, the task is completed. The movement of each agent follows a pre-defined Markov chain in the gridworld.  Using the  \ac{mdp} formulation, the state vector  is given as $\vec{X}=[X_0, X_1, X_2]$ where $X_0$ represents the position of robot and $X_1$, $X_2$ represent the positions of agent 1 and agent 2. Both agents and robot can move in four compass directions. The probabilistic transition function is defined by 
\begin{equation}
\begin{split}
&P([x'_0,x_1',x_2'] |[x_0,x_1,x_2],a ) \\
&=P(x_0'|x_0, a)\prod_{i\in \{1,2\}} P_i(x_i'| x_i).
\end{split}
\end{equation}
where $P(x_0' | x_0, a)$ is the probability that robot reaches position $x_0'$ after taking action $a$ from position $x_0$, $P_i(x_i'\mid x_i)$ is the probability that agent $i$ reaches position $x_i'$ from position $x_i$ in one step. In this experiment, the dynamics of two agents are the same: At each step, the agent moves to its neighbor cells with a uniform probability. If the agent runs into walls or hits the boundaries, it will stay in the same cell. The robot is stochastic, given an action, \ie, "N", the robot will reach the intended cell with  0.7 probability, and  the neighbors of the intended cell with 0.15 probability each. If the robot runs into walls or hits the boundaries, it will stay in the same cell. 

The capturing condition is given as follows: if the Euclidean distance $d(x_0,x_i)$ is below a threshold $\epsilon \ge0$, then the robot can capture the agent with probability $p$. In this experiment, we assign $\epsilon = 0$ and $p=1$. Once an agent is captured, the robot will receive a reward of $100$ and the agent will be removed from the environment. If the robot enters the penalty cell, it will receive a reward of $-20$. 

Assuming that the robot only tracks the movement of one agent at a time, then the attention functions are given by $\calF = \{f_1,f_2\}$ where $ 
f_1(\vec{X}) = [X_0,X_1],
$
and 
$
f_2(\vec{X}) = [X_0,X_2].
$
That is, the robot can query sensors to know its position and two agents' positions. The sensor cost  $c_i = 5$, for $ i = 0, 1, 2$.

   %At one time step, the robot can activate all its sensors to choose its attention function $f_k$ and the time duration $t$. %During this time duration $t$, the robot will only use a subset of sensors. Once this time duration $t$ ends, the robot will activate all its sensors and make attention selection based on its current state and cats' states. 

% Now, we will look into the following problem: 1) Given fixed reward weight, the expected reward under different time bound $T$. 2) Given different weights between goal-directed reward and sensor deactivating reward, what is the optimal attention selection policy for the robot to get the maximal weighted sum reward.

The set of attentional MDP includes: $M$ (The original  \ac{mdp}), $M_1$ (with $f_1(\vec{X}) = [X_0,X_1]$) and $M_2$ (with $f_2(\vec{X}) = [X_0,X_2]$). Given $M_1$, we design the state aggregation as follows: \[
D_1([X_0,X_1,X_2]|[X_0,X_1]) = \frac{1}{N_{\left[X_0, X_1\right]}}
\]
where $
N_{\left[X_0, X_1\right]} = |\{\vec{X}=[X_0, X_1, X_2] \mid X_2 \in \mathbf{X_2}\}|. $

Given the MDP $M_1$ obtained from state aggregation, we obtain the optimal attentional policy $\pi_1$. Similar procedure gives an optimal attentional policy $\pi_2$. 

We use linear scalarization method to solve Pareto optimal policy that trades off: 
\begin{itemize}
    \item The total reward of capturing both agents $R^G_T$.
    \item The total cost of information acquisition, \ie, the total  sensor deactivating rewards $R^I_T$. 
\end{itemize}
%First, we transform the information acquisition cost into a reward for the solution to be properly computed, given the fact that optimal policy is invariant under reward transformation\cite{ng1999policy}. 
%The sensor deactivating reward $C_k$ for one step could be calculated via: $C_k = \sum_{j \notin I_k}c_j$.
The weighted sum reward is given by:
\[
R^w_T(x,(\pi_k,t)) = w_1R^G_T(x,(\pi_k,t)) + w_2R^I_T(\pi_k, t),
\]
where $w_1,w_2$ are the weighting parameters, $w_1$ represents the weight of the total reward of capturing agents, $w_2$ represents the weight of the sensor deactivating reward, $w_1 + w_2 = 1$.
%Where $x$ is the current state,  $f_k$ is the attention function the robot chooses, $t$ is the duration time of attention function $f_k$. 

\begin{figure}
    \centering
    \begin{subfigure} [b]{0.45\linewidth}
    \centering
    \includegraphics[width = \linewidth]{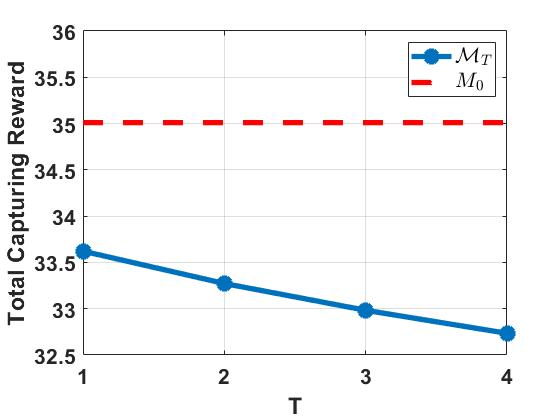}
    \caption{Total capturing reward (given different $T$).}
    \label{image_capture}
    \end{subfigure}
    \hfill
    \begin{subfigure} [b]{0.45\linewidth}
    \centering
    \includegraphics[width =\linewidth]{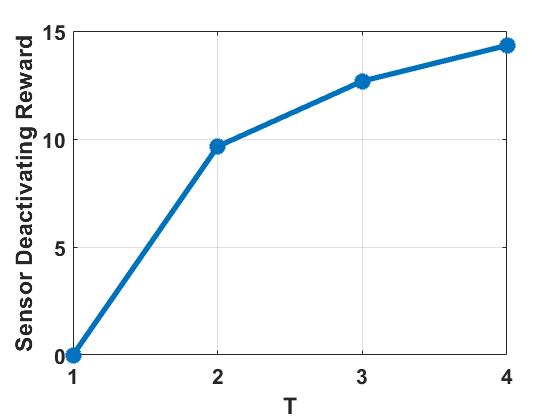}
    \caption{Sensor deactivating reward (given different $T$).}
    \label{image_sensor}
    \end{subfigure}
    \caption{The total expected capturing reward and sensor deactivating rewards  under the optimal policies in the attention shift \ac{mdp} $\calM_T$, given different upper bound $T$.}
    \label{fig: expected reward}
\end{figure}

In the first experiment, we fix the reward weight to be $[w_1,w_2]= [0.7, 0.3]$ and choose the upper bound $T = 1, 2, 3, 4$, respectively. The results are shown in Figure~\ref{fig: expected reward}. Figure~\ref{image_capture} shows the expected total capturing reward at the initial state given different upper bounds $T$, comparing to the expected total capturing reward of the initial state given the original \ac{mdp} $M_0$ which has complete observation over the states. Figure~\ref{image_sensor} shows the expected sensor deactivating reward at the initial state given different upper bounds $T$. It shows that when the time bound $T$ becomes larger, the expected sensor deactivating reward becomes larger but at a cost of decrease in the total capturing reward. As $T$ increases from $1$ to $4$, we observe the total capturing reward decreases by $2.6\%$, from $33.62$ to $32.73$. However, the sensor deactivating reward increases from $0$ to $14.6$. Also, comparing the attention shift \ac{mdp} $\calM_T$ when $T = 4$ with $M_0$, the total capturing reward at the initial state decreases by $6.5\%$, which means the attention shift \ac{mdp} does not lose much performance in the capturing task.

\begin{figure}
    \centering
    \includegraphics[width = 0.7\linewidth]{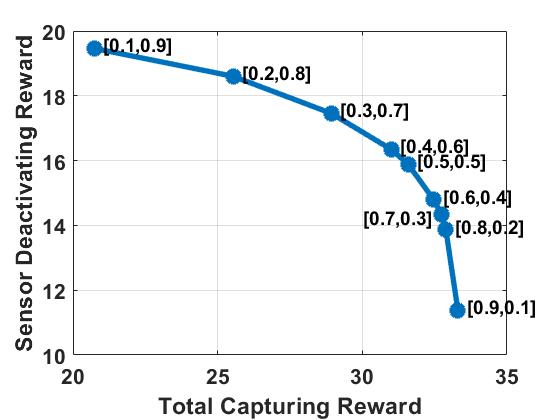}
    \caption{
    The total capturing reward and sensor deactivating reward under different reward weights, given $T = 4$.}
    \label{image_pareto}
\end{figure}

Figure~\ref{image_pareto} shows different trade-offs between the total capturing reward and total sensor deactivating reward with  different weight choices given $T = 4$. The label $[w_1, w_2]$ on the figure represents different reward weights.
From Figure~\ref{image_pareto}, we observe that as the sensor deactivating reward weight  increases, the robot increases the duration of one attention function  to reduce the cost of information acquisition. When $w_1=0.1$, the robot almost ignores its capturing task. The steep curve from $w=[0.4,0.6]$ to $w=[0.8,0.2]$ means that a large reduction in information cost can be achieved with a small loss in the total capturing reward.

\begin{figure}
    \centering
    \includegraphics[width = 0.7\linewidth]{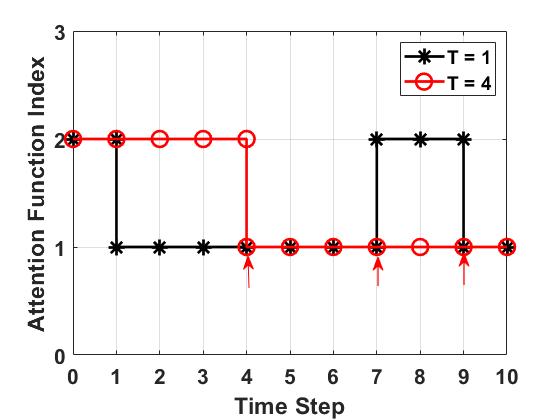}
    \caption{Attention shift process given $T = 1$ and $T = 4$.}
    \label{attention_shift}
\end{figure}

Figure~\ref{attention_shift} shows the attention shift process when $T  = 1$ and $T  = 4$ based on the same trajectory, which is obtained by running the two policies multiple times and comparing trajectories pairwise. The figure shows when $T = 1$, it may sustain an attention function longer than $T = 4$. The decision to sustain the current subpolicy is only made as the robot observes full state information at every time step. When $T = 4$, the robot only activates all its three sensors at time step $4, 7, 9$ (marked by the upper arrows in the figure). As the robot will not observe unattended states until one attention sustaining phase ends, it can save the cost of information acquisition (as reflected in Fig.~\ref{image_sensor}).

\mbox{}

%\nomenclature[01]{$b_t$}{Belief at time step $t$}
\nomenclature[02]{$f_k$}{Attention function under mode $k$}
\nomenclature[03]{$\calF$}{Set of attention function}
\nomenclature[04]{$M$}{The original MDP}
\nomenclature[05]{$M_k$}{Attentional MDP respecting attention function $f_k$}
\nomenclature[06]{$\calM_T$}{Attention-shift MDP}
\nomenclature[08]{$P$}{State transition function}
\nomenclature[09]{$R$}{Reward function}
\nomenclature[10]{$V$}{Value function of attentional \ac{mdp}}
\nomenclature[11]{$\hat V$}{Value function of attention-shift \ac{mdp}}
\nomenclature[12]{$x$}{State $x$}
\nomenclature[13]{$ X$}{State variable  X}
\nomenclature[14]{$\bm{X}$}{Set of states $x$}
\nomenclature[15{$y$}{Aggregated state $y$}
\nomenclature[16]{$\bm{Y}$}{Set of aggregated states $y$}
\nomenclature[17]{$\omega$}{Reward weighting parameter}
\nomenclature[18]{$\pi$}{Attentional policy}
\nomenclature[19]{$\Pi$}{Set of attentional policy}

\section{Conclusion}
\label{sec: conclusion}

%Understanding human's cognitive control is important to the  design of intelligent and autonomous systems. A computational model of attention control will enable us to design autonomous agents to actively select information stimuli to complete tasks under limited resources (in terms of computation power and cost of information acquisition and processing). This model is also critical to the synthesis of shared autonomous systems.  For example, a robot with an internal model of human cognitive model can  more accurately assess human's situational awareness and augment the cognitive capability of its human teammate. There are lot of works about human attention model these years.

In this work, we studied the probabilistic planning with active perception, inspired by humans' attention control ability. We  proposed an attentional \ac{mdp} with state aggregation to model the robot's attention mechanism and developed an attention-shift \ac{mdp}, in which the agent can choose its attention and the duration of the attention given the current state. We employed the linear scalarization method to find the balance between optimizing task performance and minimizing the cost of information acquisition.  As a part of the future work, we aim to extend attention-based planning for more complex task specification such as temporally evolving tasks. To reduce the computational complexity of the offline planning method, we will also investigate online planning with attention-shift. The computational model of attentions can facilitate the understanding and prediction of human's attention from observations for attention-aware human-robot collaboration.
\bibliographystyle{ieeetr}
\bibliography{refs.bib}

\end{document}